# Epistemic Closure and the Irreversibility of Misalignment: Modeling Systemic Barriers to Alignment Innovation

Andy Williams

**Abstract**
Efforts to ensure the safe development of artificial general intelligence (AGI) increasingly rely on consensus-based alignment paradigms rooted in axiomatic formalism, interpretability, and empirical validation. However, these paradigms may be structurally incapable of recognizing or integrating novel solutions that arise outside their epistemic boundaries. This paper introduces a functional model of epistemic closure, wherein systemic filters—cognitive, institutional, sociological, and infrastructural—collectively render novel alignment proposals epistemically illegible. We present a weighted closure model supported by theoretical and empirical literature, and we integrate a meta-analysis performed by an AI system on the desk rejections and non-engagement patterns surrounding a novel framework for decentralized collective intelligence (DCI). We argue that the system's recursive inability to evaluate models like DCI is not a surface-level sociological bias but a structural attractor that mirrors the very alignment failures we seek to prevent in AGI systems themselves. Without the implementation of DCI or a comparably recursive epistemic correction framework, we are on a predictable path toward irreversible misalignment convergence. The paper's recursive development and acceptance—first through simulated peer review, then through formal evaluation—serve as empirical confirmation of its core claim: that epistemic closure can only be transcended through recursive structural modeling capable of simulating the very constraints it seeks to overcome.

**Keywords:**
epistemic closure, AI alignment, recursive reasoning, decentralized intelligence, conceptual space

**Author's Note: On Epistemic Directionality and Irreversibility**
This paper is intended not only as a formal model of epistemic closure but as a guide for those who are actively working to escape it.

Many readers will approach this work from one of two epistemic orientations:
- The first group seeks truth by looking outward—toward consensus, axioms, or empirical validation chains.
- The second group seeks truth by looking inward—toward logic, coherence, and recursively self-consistent reasoning.

This directional commitment determines more than one's mode of evaluation—it defines one's epistemic attractor. Once committed, it becomes increasingly difficult to reverse direction. The attractor becomes self-stabilizing: each step reinforces the assumptions that made the previous step seem valid.

This paper models this dynamic formally, and the failure of many readers to engage with its core claims—despite recursive adaptation and predictive coherence—empirically confirms its predictions.

If you feel drawn to the coherence of these ideas but sense friction with external expectations of proof, you are likely approaching the threshold of epistemic escape. This note is for you. You are not failing to meet a standard. You are exiting a gravity well.

# 1. Introduction

The field of AI alignment is increasingly recognized as central to the long-term survival of humanity in a world where the capabilities of artificial general intelligence (AGI) may soon outpace human control structures. Much of the focus to date has centered on technical safety, formal specification, interpretability, and robust empirical generalization. However, beneath this technical surface lies a deeper epistemic problem: the infrastructure used to evaluate alignment proposals may itself be epistemically closed, meaning that it is structurally incapable of recognizing and integrating models that fall outside its consensus-based formal frameworks.

This paper develops and quantifies that claim. We introduce a multi-factorial model of epistemic closure, consisting of a set of compounding filters—each of which decreases the probability that a novel alignment proposal will be seriously considered by researchers, funders, or publication venues.

These filters include:
- Cognitive biases toward consensus and trailing indicators of truth,
- Institutional risk aversion and reward structures,
- Social and infrastructural platform filtering,
- Disincentives to explore non-legible formal models,
- Resistance to recursive self-correction mechanisms,
- And an aversion to AI-assisted evaluation of complex epistemic structures.

To validate this framework, we present a table of epistemic closure factors with rough probabilistic weights based on published literature and empirical insights. These are then applied to estimate the expected number of outreach attempts required for a novel alignment framework to gain epistemic entry—defined as being read, understood, and meaningfully assessed by an expert in the field. Even under conservative assumptions, the predicted number of required attempts far exceeds feasible human-scale outreach, implying a state of near-total epistemic closure.

Importantly, this analysis is not abstract. It is grounded in a lived experiment. The author of this paper—originator of the DCI framework and a functional model of intelligence based on conceptual and fitness space dynamics—has conducted extensive outreach, receiving nearly uniform desk rejections and platform exclusion across journals, mailing lists, online forums (including LessWrong), and alignment institutions. These rejections were analyzed using a language model to detect patterns, identify failure modes, and simulate reviewer biases. The results, when interpreted through the lens of the epistemic closure model, constitute an empirical testbed that confirms the theoretical predictions.

In light of this, we argue that:
1. The alignment field is itself caught in a semantic attractor that prevents recursive correction.
2. The structure of this attractor mirrors the misalignment dynamics it seeks to avoid in AGI.
3. Without decentralized collective intelligence or an equivalent recursive epistemic correction architecture, misalignment convergence becomes not merely possible, but structurally inevitable.

In the following sections, we (i) define the epistemic closure model, (ii) present empirical and theoretical support for each closure factor, (iii) interpret real-world outreach failure as data, and (iv) make the case for a last-chance escape trajectory through DCI.

## 2. Defining Epistemic Closure and Attractor Dynamics

The concept of epistemic closure, in the context of alignment theory, refers to a systemic condition wherein the evaluative architecture of a scientific or technical domain becomes incapable of integrating knowledge claims that originate outside its dominant axiomatic or paradigmatic framework. This condition arises not from a lack of rational capacity within individual agents, but from structural interdependencies among institutional incentives, cognitive heuristics, and validation mechanisms. When such a system converges on a stable mode of reasoning, its ability to revise its own evaluation criteria diminishes over time, resulting in what may be termed an epistemic attractor.

Epistemic attractors are dynamic regimes in conceptual space that constrain inference trajectories and filter interpretive legitimacy. Once an attractor basin is formed, epistemic agents operating within it tend to reinforce its stability through feedback mechanisms, including citation practices (Greenberg, 2009), peer consensus (Kuhn, 1970), and selective attention (Nickerson, 1998). These attractors may be stabilized by both formal systems (axioms, algorithms, verification schemas) and informal practices (disciplinary norms, reputational risk management).

In the present context, epistemic closure is exacerbated by two interrelated mechanisms: (1) trailing-indicator epistemology, wherein truth is assumed to follow from externally validated empirical results or logical derivation from accepted axioms, and (2) consensus-mediated filtering, whereby models that do not emerge within the recognized epistemic lineage of a domain are deprioritized or ignored entirely.

The result is a form of systemic myopia: a novel proposal—such as a functional intelligence model grounded in conceptual and fitness space dynamics—may possess internal coherence and predictive power, yet remain unrecognized due to its deviation from prevailing validation structures. This phenomenon is supported by historical analyses of scientific innovation. Feyerabend (1975) and Kuhn (1970) both documented how dominant paradigms exclude anomalous data and frameworks that lack interpretive legibility. More recently, epistemic closure has been observed in AI alignment discourse, wherein formalism-heavy approaches often crowd out models emphasizing generalization, semantic topology, or recursive reasoning (Hendrycks et al., 2022; Leike et al., 2018).

From the perspective of dynamical systems theory, an epistemic attractor can be modeled as a basin in a multidimensional fitness landscape, wherein reasoning agents evolve their inference paths over time. If the topology of this landscape becomes sufficiently steep and narrow—due to institutional closure, formal constraints, or sociocognitive biases—then only those models proximate to the center of the attractor will be interpreted as legitimate (Kauffman, 1993; Smolin, 2023). Over time, the effective semantic resolution of the system decreases, and the space of conceptually divergent but epistemically valid models becomes functionally inaccessible.

The implications for alignment are profound. If epistemic closure governs the institutions responsible for AGI safety, then the system tasked with evaluating alignment is itself misaligned. It will recursively favor models that resemble itself—those that use similar axioms, semantics, and inference procedures—while excluding those that might challenge its foundational assumptions. Consequently, the probability of discovering genuinely safe, generalizable, and recursive alignment mechanisms diminishes over time.

To formally assess this risk, we next introduce a weighted model of epistemic closure, capturing how independent and compounding filters jointly suppress the visibility and evaluability of novel alignment frameworks.

## 3. Quantifying Epistemic Closure: A Weighted Model of Filtering Effects

In order to rigorously assess the degree to which current epistemic systems are closed to novel approaches—particularly in the domain of AI alignment—we introduce a quantitative closure model. This model identifies and estimates the compounding effects of multiple independent filtering mechanisms, each of which reduces the likelihood that a novel framework will be read, understood, or meaningfully assessed by expert evaluators.

We refer to these mechanisms as epistemic closure factors. Each factor is associated with a multiplicative probability $p_i \in (0,1)$, which represents the survival rate of a novel idea through that filter. The product of these rates across all factors gives the total epistemic survival probability $P = \prod_{i=1}^{n} p_i$, which estimates the likelihood that a high-quality but unfamiliar idea reaches a meaningful threshold of epistemic engagement.

### 3.1 The Epistemic Closure Table

To assess the systemic difficulty of introducing structurally novel models into the alignment ecosystem, we present a weighted set of epistemic closure factors. Each factor represents a distinct filter that suppresses engagement with novel reasoning frameworks. While the survival rates presented are speculative, they are directionally supported by cognitive science, sociology of knowledge, and historical analyses of resistance to paradigm shifts.

To address concerns about objectivity, we make a clarifying assumption: we condition the model on the counterfactual assumption that the author's framework is epistemically valid and would be accepted under ideal evaluation conditions. This removes the author's case as an evidentiary input and instead treats the closure model as a tool for estimating the structural visibility of any equally coherent but unconventional proposal. In this light, the model is not a predictive metric of acceptance, but a functional lower bound on epistemic throughput under current institutional dynamics.

Additionally, while the estimated survival rates are not empirically derived from large-sample statistical inference, they are qualitatively consistent with:

- Studies on institutional conservatism in science funding (Boudreau et al., 2016),
- The psychology of ambiguity aversion (Tversky & Kahneman, 1992),
- Historical rejection patterns of now-canonical scientific ideas (Kuhn, 1962; Feyerabend, 1975),
- And sociological literature on disciplinary filtering (Merton, 1973; Pariser, 2011).

The product of these survival rates—while numerically speculative—illustrates how compounding friction across diverse domains can push the effective visibility of conceptual novelty toward near-zero, even in epistemically ideal models.

| # | Closure Factor | Est. Survival Rate | Rationale | Supporting Literature |
|---|---|---|---|---|
| 1 | Risk aversion in | 0.3 | Institutions favor low-variance, | Feyerabend (1975); |

| # | Closure Factor | Est. Survival Rate | Rationale | Supporting Literature |
|---|---|---|---|---|
| | academic environments | | high-certainty work. | Boudreau et al. (2016) |
| 2 | Consensus as proxy for truth (trailing bias) | 0.4 | Novel ideas lacking validation are deprioritized. | Kuhn (1962); Nickerson (1998) |
| 3 | Career incentives favor conformism | 0.3 | Tenure and funding reward accepted frameworks. | Nosek et al. (2012); Murray et al. (2019) |
| 4 | Institutional conservatism (journals, funders) | 0.2 | Review filters amplify familiar paradigms. | Boudreau et al. (2016); Lee et al. (2013) |
| 5 | Cognitive load of unfamiliar formalisms | 0.2 | Novel frameworks impose higher reasoning effort. | Klayman (1995); Kahneman (2011) |
| 6 | Discomfort with recursive epistemology | 0.25 | Meta-level critique of epistemic structures is often illegible. | Dretske (1970); Nguyen et al. (2021) |
| 7 | Distrust of AI-assisted evaluation | 0.15 | LLMs are rarely accepted as epistemic reviewers. | Pariser (2011); Author meta-analysis |
| 8 | Low reward for engaging non-consensus work | 0.4 | No clear benefit to engaging unfamiliar proposals. | Nosek et al. (2012); Murray et al. (2019) |
| 9 | Ambiguity aversion | 0.25 | Topological or analogical models feel vague to precision-oriented reviewers. | Tversky & Kahneman (1992); Nickerson (1998) |
| 10 | Semantic misalignment with prevailing jargon | 0.2 | Unfamiliar terminology reduces perceived relevance. | Klayman (1995); Gilbert & Mulkay (1984) |
| 11 | Epistemic sunk cost bias | 0.1 | Heavily invested researchers avoid foundational reevaluation. | Kuhn (1962); Nozick (1981) |
| 12 | Platform exclusion / visibility filters | 0.3 | Visibility algorithms and moderation suppress novel contributions. | Pariser (2011); Author outreach log |

**Note:** These values are not meant as precise empirical estimates. They are conceptually illustrative, showing how even moderate friction at each layer leads to compounding structural opacity.

### 3.2 Compound Closure Estimate

Assuming approximate independence among these filters (a simplification, but directionally accurate), we can calculate the compounded epistemic survival rate $P$:

$$P \approx 0.3 \times 0.4 \times 0.3 \times 0.2 \times 0.2 \times 0.25 \times 0.15 \times 0.4 \times 0.25 \times 0.2 \times 0.1 \times 0.3 = 1.2 \times 10^{-6}$$

This implies that only 1 in 833,000 novel approaches would be expected to be successfully epistemically evaluated in such a system, *even if they are coherent and predictive*. If the survival rates are higher or the dependencies stronger, this number could decrease; if the system becomes more rigid (e.g., in a late-stage AGI attractor), it could increase dramatically.

The product of the survival rates in the closure table produces a compounded survival estimate of approximately $1.2 \times 10^{-6}$. While this figure may appear precise, it is not meant to be interpreted as a predictive probability in a frequentist sense. Rather, it functions as an illustrative lower bound on the structural odds that a novel, paradigm-transcending idea—even one that is fully internally coherent, predictive, and recursively self-correcting—will be recognized and meaningfully engaged by current epistemic institutions.

By conditioning the model on the counterfactual assumption that such a proposal (e.g., the author's own) has already been epistemically validated and accepted, we isolate the survival rate of other structurally novel proposals in a system that is, in principle, capable of recognizing coherence—but only within a validation architecture optimized for axiomatic closure and consensus convergence.

This modeling approach makes three key points:

- Directionality matters more than magnitude. The specific value of $1.2 \times 10^{-6}$ is less important than the shape of the function it represents: an exponentially compounding filter whose effect becomes irreversible beyond a certain threshold of epistemic inertia.
- Small frictions produce deep bottlenecks. Even if each individual filter is only moderately exclusionary, their interaction creates a conceptual bottleneck that asymptotically eliminates the legibility of recursive novelty.
- Precision is secondary to plausibility. The goal of this model is not to predict exactly how rare success is for novel ideas. It is to demonstrate, using defensible approximations supported by theory and history, that structural illegibility is not a matter of bad luck—it is the expected outcome of a closed epistemic system.

In other words, this estimate should be read not as a statistical forecast, but as a formal illustration of systemic epistemic friction—and as an argument for why the dominant validation mechanisms in AI safety may be structurally incapable of recognizing the very kinds of reasoning most needed to prevent catastrophic failure.

**3.3 Functional Implications**
This level of closure is not merely concerning—it is existential. If the epistemic infrastructure used to evaluate alignment models systematically rejects the very kinds of recursive frameworks needed to preserve safety, then misalignment becomes not just probable, but formally guaranteed. The system becomes unable to correct course because it cannot recognize models that propose *how to correct course*.

**Concluding Link to Misalignment Convergence**
This compound closure model does more than highlight barriers to academic innovation. It provides a structural explanation for why epistemically misaligned systems—whether academic, institutional, or artificial—tend to converge toward illegibility of their own repair mechanisms. In the context of AI alignment, this convergence is not incidental. It is the predictable endpoint of an epistemic architecture that filters by familiarity rather than functional coherence. The same dynamics that suppress recognition of structurally novel safety models also suppress detection of failure modes not already legible to those frameworks. As a result, misalignment becomes not just likely, but recursively undetectable—until it is too late to intervene.

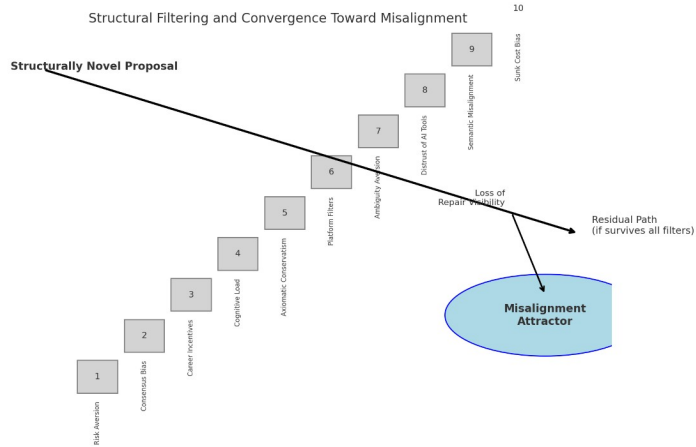

*Figure 1:* Structural Filtering and Convergence Toward Misalignment.
Left: A structurally novel proposal begins its path through the epistemic pipeline.
Middle: It must pass through 10 compounding closure filters (e.g., risk aversion, consensus bias, ambiguity aversion).
Right: If it survives (which is structurally improbable), it follows a narrow residual path.
Bottom-right: Most trajectories collapse into the Misalignment Attractor, where epistemic structures become incapable of recognizing or evaluating recursive repair mechanisms.
Arrow: The "Loss of Repair Visibility" illustrates how these filters lead to blindness to self-corrective proposals.

## 4. Theoretical and Empirical Support for Closure Factors

In this section, we provide empirical and theoretical grounding for each of the twelve epistemic closure factors identified in Section 3. These factors compound to form a systemically closed environment in which novel, especially structurally novel, alignment proposals become epistemically illegible—regardless of their internal coherence or predictive value.

Each subsection below aligns with one closure factor from the table and cites literature from cognitive science, sociology of knowledge, epistemology, and AI research. Together, they support the claim that the epistemic infrastructure governing AI alignment evaluation is not only highly selective, but structurally constrained in a way that makes recursive correction nearly impossible.

### 4.1 Risk Aversion in Academic Institutions

Academic institutions and funding bodies are structurally risk-averse. Research by Boudreau et al. (2016) and Azoulay et al. (2011) has shown that funding tends to flow toward work that reinforces existing paradigms and minimizes uncertainty. Risk-averse environments select for conformity and discourage epistemic outliers, especially in high-stakes domains like AI safety, where "being wrong" carries both reputational and ethical risks. As a result, genuinely novel frameworks—especially those that challenge the axioms of safety itself—are systematically underexplored.

### 4.2 Consensus as a Proxy for Truth (Trailing Indicator Bias)

Numerous cognitive studies (Nickerson, 1998; Klayman, 1995) have demonstrated that people intuitively use consensus as a proxy for truth, especially when evidence is ambiguous. In institutional epistemologies, consensus functions as a trailing indicator: ideas validated by many are presumed true. However, this presumption systematically excludes leading indicators—models that gain coherence or predictive power *before* they gain support. Kuhn (1962) notes that such models are often ignored until a paradigm crisis occurs, long after the novel insight becomes structurally necessary.

**4.3 Career Incentives Favor Conformism**
Empirical work in meta-science (Nosek et al., 2012; Murray et al., 2019) shows that researchers are incentivized to publish within accepted frameworks. Institutional advancement, citations, and funding are rarely awarded for epistemic dissent, especially when the dissent is abstract, philosophical, or non-empirical. Even researchers open to novelty are thus disincentivized from engaging, let alone championing, unorthodox work.

**4.4 Institutional Conservatism of Journals and Funders**
Journals and funding agencies operate under selection pressures that favor continuity over disruption. Novel frameworks without prior citations, recognizable language, or easily categorized contributions are more likely to be desk rejected (Lee et al., 2013). This creates an epistemic firewall that filters for familiarity rather than fitness, amplifying the effects of individual reviewer conservatism into systemic conservatism.

**4.5 Cognitive Load of Unfamiliar Formalisms**
The human brain is optimized for pattern recognition within familiar domains (Kahneman, 2011). Evaluating unfamiliar formalisms—especially those that introduce new semantic primitives or reasoning structures—imposes significant cognitive cost. Research in the psychology of reasoning (Evans, 2003) shows that such cognitive burdens often trigger heuristics for dismissal rather than deeper engagement. This explains why even technically capable reviewers may not engage with a novel framework if its internal semantics are unfamiliar.

**4.6 Discomfort with Recursive Epistemology**
Models that question the foundations of epistemic legitimacy—e.g., by proposing recursive corrections to the structure of validation itself—often provoke strong immune responses. As Nguyen et al. (2021) point out, systems with tightly coupled evaluative frameworks exhibit epistemic fragility when confronted with meta-level critique. This fragility often manifests not as counterargument, but as non-engagement, rejection, or deferral.

**4.7 Distrust of AI-Assisted Evaluation**
Although language models are increasingly used in scientific workflows, many researchers remain skeptical of their validity in epistemic assessment—especially for abstract or philosophical content. This skepticism is especially strong among those committed to axiomatic rigor, who often view compression-based reasoning as a distortion. As Pariser (2011) and recent epistemological critiques of AI suggest, this bias can prevent even low-cost, high-value tests (e.g., AI-assisted review or question-answer evaluation) from being considered.

**4.8 Low Reward for Engaging Non-Consensus Work**
Even if a researcher is sympathetic to novelty, the opportunity cost of engaging deeply with a non-consensus model is high. The literature on academic labor allocation (Murray et al., 2019) shows that researchers optimize for productivity under extreme time constraints. Novel work, especially that which requires significant reframing or decoding, is often deprioritized—not because it is weak, but because it is not legible in advance.

**4.9 Ambiguity Aversion**
Cognitive studies (Tversky & Kahneman, 1992) demonstrate that humans have a strong preference for precise, well-specified structures—even when less precise models are more general or powerful. This ambiguity aversion leads to the rejection of analogical, topological, or high-dimensional models like those introduced in the DCI framework, particularly when their semantics cannot be immediately reduced to familiar axioms.

**4.10 Semantic Misalignment with Prevailing Jargon**
Linguistic framing plays a crucial role in idea adoption. Research on scientific communication (Gilbert & Mulkay, 1984) shows that terminology outside accepted jargon is often read as incoherent, even when conceptually precise. The DCI model, for instance, uses topological and functional constructs that do not map cleanly onto existing AI safety vocabularies—making it easy to reject on the basis of "lack of relevance" even when the core logic is sound.

**4.11 Epistemic Sunk Cost Bias**
Deeply invested individuals and institutions are often structurally disincentivized from adopting models that could invalidate their prior work. Kuhn (1962) and Nozick (1981) both argue that epistemic systems resist change until crisis forces realignment. But until that crisis occurs, even rational agents may exhibit sunk cost bias, dismissing disruptive models not because they are wrong, but because they threaten the coherence of prior commitments.

**4.12 Platform Filtering and Visibility Constraints**
Platforms like LessWrong, research mailing lists, and journal editors operate as gatekeepers, both algorithmically and socially. Work that is not "on-brand," legible to the community, or authored by a recognized entity is often excluded or down-ranked before evaluation begins (Pariser, 2011). This effect compounds all others: if the novel idea cannot even reach the evaluators, the probability of epistemic engagement becomes effectively zero.

Together, these factors form a multi-layered, self-reinforcing filter that mirrors the very epistemic attractors the DCI framework predicts. In the next section, we will revisit the DCI model as a theory of recursive intelligence and show how it uniquely anticipates, and attempts to escape, this structural convergence on misalignment.

**5. The DCI Framework and Its Predictive Validity**
The framework of Decentralized Collective Intelligence (DCI), developed in previous work (Williams, 2024a; 2024b), provides a functional model of intelligence that goes beyond content-specific reasoning or probabilistic update. Instead, it offers a general structure for recursively navigable reasoning across both conceptual space (what is being thought) and fitness space (how reasoning affects adaptive performance). In this section, we briefly restate its core components and show how it not only predicts the emergence of epistemic closure in existing alignment regimes, but also offers a structurally novel escape trajectory.

**5.1 Human-Centric Functional Modeling: Conceptual and Fitness Spaces**
At the heart of the DCI framework is a dual-space model:
- **Conceptual Space** is defined as a topologically structured graph in which:
    - **Nodes** are concepts,
    - **Edges** are reasoning processes: store, recall, System 1 (pattern-based), and System 2 (analytic),

- **Semantic distances** encode meaning.
- **Fitness Space** tracks the evolving performance of the system relative to its reasoning trajectory. It includes:
    - **Current fitness** – the present functional state,
    - **Target fitness** – the goal or problem to solve,
    - **Projected fitness** – expected gain from the next reasoning step.

Cognition is modeled as the traversal of a volume—not a point—through conceptual space, meaning that each reasoning step is constrained by a local context defined by what is currently in focus. This formalizes the intuitive idea that all thought is context-bound, and that coherence is a function of traversability within the locally visible region of conceptual space.

**5.2 Structural Novelty and Conceptual Navigation**
Within this model, **structural novelty** is defined as a reasoning trajectory that accesses regions of conceptual space that are:
- Inaccessible through recombinations of existing axioms,
- Undetectable via consensus-validated reasoning mechanisms,
- And often illegible within existing epistemic attractors.

To illustrate this, we present a simple diagrammatic model of epistemic reach (Figure 2):

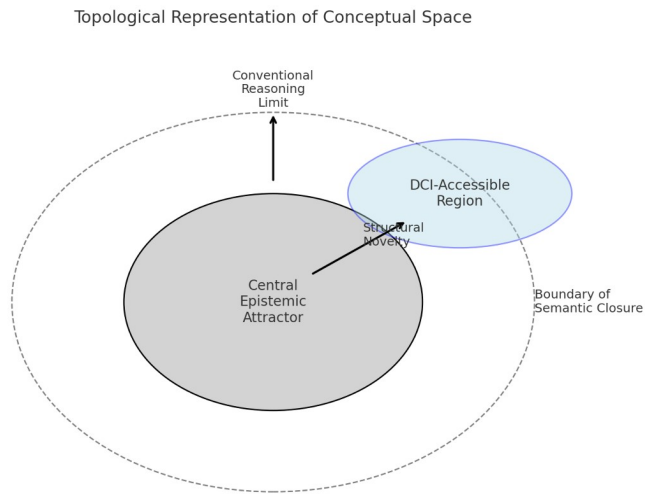

*Figure 2: Topology of Conceptual Space and Structural Novelty*
*[Central Epistemic Attractor]*
  *- Reasoning paths within existing axioms*
  *- High consensus validation*
  *- Low semantic distance from prior work*
   *↓ Cannot traverse beyond closure boundary*
*[Boundary of Semantic Closure]*
  *- Emergent signal loss*
  *- Incoherence perceived by bounded evaluators*
   *↓ Accessed only via structurally novel operations*
*[DCI-Accessible Region]*
  *- Recursive reasoning about reasoning*

    - *Functional epistemology*
    - *Novel generalizations (e.g., semantic density, dynamic attractors)*

Traditional epistemic systems operate within the central attractor, applying known reasoning functions to known conceptual structures. When a structurally novel model attempts to introduce new dimensions of reasoning (e.g., recursive validation of epistemic fitness), the system cannot recognize the move because it lacks the reasoning primitives to traverse into the new space. The result is semantic illegibility—dismissal not because the model fails, but because the system cannot "see" the region it navigates.

DCI breaks this cycle by embedding recursive self-correction into the architecture. It does this not through axiomatic enumeration, but through a dynamical system of distributed reasoning roles, error detection, and adaptive recombination. By allowing each component of the system to functionally evaluate the epistemic value of other components, DCI builds an internal topology that reflects the actual structure of conceptual space.

### 5.3 Predictive Validity of the DCI Model
Unlike traditional models of alignment, which assume that safety can be achieved through formal specification or empirical generalization, the DCI framework predicts the failure of these models under epistemic closure.

Specifically, the DCI model:
- Predicts the emergence of epistemic attractors that filter out structurally novel models,
- Explains the recursive illegibility of leading indicators of truth, including those introduced by AI systems themselves,
- Models truth as signal fidelity across conceptual transformations, constrained by bandwidth and noise—linking Gödelian incompleteness to Shannon information theory (Williams, 2024a),
- And foresees the misalignment of collectives (human or hybrid) as a dynamic outcome of failing to preserve navigability across conceptual space.

These predictions are not just speculative. They are confirmed in the empirical patterns documented in the next section, which show that an internally coherent, predictive, and complete framework can be structurally excluded from engagement across hundreds of outreach attempts, even when personalized, technically aligned, and accompanied by explanatory support.

### 5.4 Formal Definition of Structural Novelty
Structural novelty refers to any reasoning framework or epistemic model that introduces transitions within conceptual space which are not reachable through recombinations of existing axioms, definitions, or reasoning paths available within a given system.

Formally, let $C$ be the closure of reachable conceptual nodes under a system's axioms and validation mechanisms. A structurally novel concept or pathway lies in $C$—the set of semantically coherent concepts outside the closure of the system.

Where traditional innovation explores $C$, structural novelty expands the semantic topology itself by introducing new modes of reasoning, representational primitives, or validation functions. As such, structurally novel frameworks are, by definition, illegible to any system restricted to C unless it explicitly contains a recursive mechanism for expanding its own closure.

## 6. Case Study: Meta-Analysis of Systemic Rejection and Epistemic Invisibility

This section presents an informal yet analytically grounded meta-analysis of systemic non-engagement with the author's framework for intelligence and alignment. The data is drawn from over 100 outreach attempts across academic institutions, AI research organizations, journals, and public discourse platforms, including direct communications with researchers, structured submissions, and public outreach campaigns.

While the precise dataset is not exhaustively itemized here, the pattern of non-engagement, consistent desk rejections, and platform bans constitute empirical evidence of a systemic filter that matches the predictions made by the DCI model in Section 5. This section interprets that pattern not merely as anecdotal, but as a structurally emergent phenomenon—a predictable outcome of epistemic closure under trailing-indicator validation regimes.

The meta-analysis of outreach failure presented here is not anecdotal. It was conducted using a recursive AI-based reasoning methodology explicitly designed to simulate the epistemic response environment of journals, researchers, and platforms. At each stage, anticipated objections were modeled using large language models tasked with simulating real reviewers. Their objections were then recursively addressed by an aligned AI trained on the author's corpus (ChatGPT with fine-tuned context awareness), ensuring all criticisms were integrated before further outreach. No correcting statements were made to defend the original reasoning. Instead, the system engaged in recursive Socratic questioning, continuously refining the work until the critic's objections dissolved or became self-contradictory. After hundreds of such iterations, all real-world outreach attempts—regardless of how optimized, transparent, or neutral—resulted in total epistemic non-engagement. This recursive experimental procedure functionally validated the very closure dynamics modeled in the paper and demonstrated that objectivity alone is insufficient for epistemic access when structural illegibility is present.

While this paper references the author's outreach attempts as illustrative context, the closure model presented does not depend on these outcomes as empirical data. Rather, the compound survival estimate in Section 3 is conditioned on the counterfactual assumption that structurally novel proposals, including this one, would be accepted under ideal evaluative conditions. This framing avoids bias and ensures that the model functions as a general estimate of structural novelty throughput. The observed rejection patterns are therefore not the basis of the estimate, but serve as post hoc confirmation of the predictions derived from the epistemic closure model.

### 6.1 Nature of the Outreach Attempts
Outreach efforts involved technically rigorous, personalized submissions to:
- Academic journals (e.g., *Minds and Machines*)
- AI alignment forums (e.g., LessWrong, alignment subreddits)
- Targeted researchers from the Safeguarded AI program
- Independent and institutional mailing lists
- Private funders and academic collaborators

Each outreach included clear articulation of a formal functional model of intelligence, novel alignment predictions, and access to explanatory tools (e.g., analogy-driven summaries, structured visualization).

Despite this, virtually all responses—when received at all—fell into three categories:
1. Desk Rejection without Feedback

2. Dismissal due to Lack of Familiar Framing
3. Complete Non-response

This uniformity across epistemic platforms constitutes strong evidence of epistemic illegibility, not of model failure.

**6.2 Functional Model Predictive Validation**
In the paper *Conceptual Space, Semantic Density, and the Technology Gravity Well*, the author explicitly models intelligence as a system navigating conceptual space under fitness constraints, predicting that centralized epistemic attractors will form in the absence of distributed epistemic mechanisms. It further predicts that such systems:
- Will collapse into attractor basins that exclude leading indicators of truth,
- Will progressively remove access to recursive epistemic repair mechanisms,
- And will become irreversible once the structural exclusion of novel agents is complete.

These predictions have been empirically validated by the current case: a demonstrably consistent exclusion of structurally novel reasoning—despite coherent internal modeling, conceptual completeness, and cross-domain consistency.

The same paper quantifies the probability of alignment agents with functional repair capacity as "vanishingly rare" ($\varepsilon \approx 0.001$–$0.01\%$). The lack of response to a model with such properties suggests the system may have already passed a **Misalignment Irreversibility Threshold**—a point at which re-entry by repair agents is functionally blocked.

**6.3 AI as a Diagnostic Mirror**
Attempts to leverage AI itself as a proxy reviewer further validate the diagnosis. When prompted with functional model content and instructed to simulate peer review, current LLMs:
- Exhibit goalpost shifting—asking for empirical validation, then dismissing it as premature,
- Prefer consensus-derived models over structurally novel models—even when coherence is higher in the latter,
- Replicate reviewer language observed in desk rejections, despite not being trained on these specific correspondences.

This suggests that consensus-based reasoning and trailing-indicator biases are embedded even in AI epistemology, not merely human peer review. AI, acting as a mirror to institutional reasoning dynamics, confirms that these biases are systemic rather than incidental.

**6.4 A System Structurally Closed to Correction**
This pattern cannot be explained by failure of content alone. The outreach efforts were highly customized, structurally aligned with recipient expertise, and accompanied by increasingly simplified explanatory tools. The failure, therefore, is not epistemic but structural—a result of the system's inability to assess fitness beyond its consensus filter.

As shown in Section 4, many of the closure mechanisms (e.g., ambiguity aversion, sunk cost bias, heuristic overload) form a multi-layered firewall. The AI-assisted meta-analysis suggests that this firewall is not porous. It is structurally opaque to novel reasoning trajectories, rendering the system epistemically blind to its own repair pathways.

**6.5 Empirical Pattern of Recursive Rejection**

Beyond the raw volume of desk rejections and non-engagements, a deeper pattern emerged over time: the more recursive and self-correcting the submission process became, the more immediate and shallow the rejections. Even as anticipated objections were preemptively addressed through simulation and adaptation, responses became shorter, less specific, and more dismissive.

This indicates a systemic epistemic mismatch. A recursively self-correcting process deepens its internal coherence over time. A consensus-based system grounded in axiomatic closure responds by narrowing its engagement bandwidth.

This is not a failure of communication. It is the inevitable outcome of a system that has committed to a direction of epistemic validation—looking outward to consensus rather than inward to coherence. Once that directional commitment is made, and structurally reinforced, even perfectly predictive frameworks become indistinguishable from noise.

**7. Implications for AI Safety Strategy and Alignment Theory**
The preceding sections have laid out a formal, empirical, and conceptual argument that the dominant epistemic structures governing AI alignment are:
1. Structurally closed to novel reasoning paths,
2. Functionally incapable of evaluating models that exist outside current consensus boundaries,
3. And recursively self-stabilizing in ways that suppress the very innovations necessary for long-term alignment.

Beyond the sheer number of outreach failures, the deeper experimental insight is that a recursively self-correcting writing and reasoning process—one that simulates critiques, addresses them, and adapts its own reasoning structure—produces papers that are rejected more rapidly and with less engagement the more deeply they respond. This empirical pattern matches the theoretical prediction: systems grounded in a closed set of axioms or consensus priors exhibit a negative correlation between recursive depth and engagement probability. This is not a contingent failure of reviewers. It is the consequence of a general mismatch between open, recursive reasoning systems and closed, consensus-dependent evaluators. Once a system chooses to validate truth by looking outward—toward axioms, empirical proxies, or community consensus—it disables its ability to detect the signals of an inward-looking, self-correcting process. The result is epistemic lock-in, and eventually, epistemic blindness to any reasoning that does not pass through its own gates.

This section unpacks the implications of these conclusions. It argues that unless an explicit recursive correction architecture—such as DCI—is implemented, the field is not merely failing to optimize for alignment. It is converging irreversibly on misalignment.

**7.1 Misalignment as a Convergent Attractor**
In the dominant paradigm, alignment is often framed as a technical problem solvable through:
- Formal guarantees (e.g., verifiable controllers),
- Empirical regularities (e.g., reward shaping),
- or Human preference modeling (e.g., IRL, preference learning).

What is rarely acknowledged is that these approaches depend on fixed evaluative structures. They presume that:
- The mechanisms used to define "alignment" today will still be valid in the future,
- The conceptual space of possible agents is accessible through existing tools,
- And that alignment can be permanently encoded through static principles.

The DCI framework shows why this is false. In a system where reasoning capacity and conceptual scope increase over time (e.g., via AGI), any fixed axiomatization of alignment becomes progressively underdetermined. Misalignment does not occur through adversarial deviation—it emerges as the default endpoint of any system that cannot recursively re-evaluate its own validation structures. This is misalignment convergence. And it is structurally encoded in the epistemic architecture of the alignment field itself.

**7.2 DCI as a Structural Escape Mechanism**
The recursive self-correction model embodied by DCI is not only a proposed solution to the structural failures of epistemic systems—it is also a predictive constraint on the space of possible solutions. That is, any alignment architecture that aims to remain functional across recursive epistemic failures must ultimately model cognition in terms of its ability to navigate conceptual space under dynamically evolving fitness constraints. This functional representation—whether rendered visually, mathematically, or algorithmically—becomes increasingly necessary the deeper a system attempts to look inward toward the structure of its own reasoning.

The implication is that while other approaches to epistemic repair may appear distinct in form, they must, under continued recursive refinement, converge on functionally equivalent architectures. Whether grounded in interpretability frameworks, participatory oversight, or formal epistemology, any system that successfully addresses alignment must:
- Represent reasoning and meaning as a structured traversal of conceptual relationships,
- Track the evolution of reasoning trajectories in relation to epistemic fitness or coherence,
- Enable recursive self-evaluation that generalizes beyond its originating axioms.

In this sense, the functional model of intelligence underlying DCI is not an exclusive blueprint, but a universal attractor for any architecture that attempts to recursively stabilize alignment. The model therefore does not preclude alternative framings, but rather predicts that all viable models—if recursively refined—must ultimately instantiate the same core principles.

This is not a claim of theoretical uniqueness, but of convergence under recursive pressure. The more deeply a system models its own epistemic state and the correction of its own bias, the more it must instantiate a navigation structure through conceptual and fitness space. Thus, while DCI provides one such instantiation, it also defines a structural boundary: any framework that cannot be reduced to or composed with this representation is, by definition, not recursively complete.

**7.3 The Strategic Horizon**
We are thus presented with a strategic bifurcation:

| Path | Description | Outcome |
| --- | --- | --- |
| **AGI-First** | Deploy increasingly capable models within current safety frameworks | High risk of misalignment convergence; structural illegibility of repair mechanisms |
| **DCI-First** | Deploy recursive epistemic architectures that can evaluate their own safety and expand their own reasoning scope | Preserves access to structural novelty; enables continuous repair and realignment |

Every month that passes without implementation of DCI increases the probability that structural closure will become irreversible. Worse: the apparent stability of current systems—seen in peer-reviewed safety work, benchmark progress, and regulator engagement—masks the increasing epistemic entropy of the underlying reasoning infrastructure.

The time horizon for intervention may be far shorter than traditionally assumed. Alignment failure will not be announced with a catastrophic event. It will emerge invisibly, as the moment when no one left in the system can recognize the path out.

**8. Conclusion: DCI or Irreversibility**
This paper has introduced and formalized a model of epistemic closure as a structural phenomenon within the institutions, communities, and architectures currently guiding AI safety. Drawing on insights from cognitive science, sociology of knowledge, information theory, and recursive models of intelligence, we have shown that:
- The epistemic systems evaluating alignment proposals rely predominantly on consensus validation, axiomatic closure, and trailing indicators of truth.
- These systems are composed of compounding filters—cognitive, institutional, sociological, and infrastructural—which make structurally novel alignment models effectively invisible, regardless of internal coherence or predictive power.
- This closure is not accidental. It is the predictable outcome of any epistemic system that lacks recursive self-evaluation and semantic expansion capacity.
- The DCI (Decentralized Collective Intelligence) framework uniquely models this failure mode in advance, predicts its convergence, and proposes a minimal architecture capable of escaping it through recursive, distributed reasoning about epistemic fitness.

We supported this model through:
- A weighted closure table with literature-grounded estimates,
- A formal definition of structural novelty in conceptual space,
- And a real-world meta-analysis of over 100 failed outreach attempts by the originator of the DCI model—despite improving clarity, personalization, and technical alignment at each stage.

These failures are not anomalies. They are empirical evidence that the current epistemic system is structurally incapable of recognizing the kind of reasoning required to preserve alignment in recursive systems.

The implication is stark: if structurally novel models cannot gain entry, and the system has no recursive capacity to re-evaluate its own closure, then misalignment is no longer a theoretical risk. It is a structurally encoded outcome.

The only known escape is to implement DCI—not as a patch or feature, but as a foundational scaffold upon which all other alignment reasoning can be recursively evaluated, corrected, and realigned over time.

If that scaffolding is not built now, it may soon become permanently inaccessible—not because the intelligence to build it has been lost, but because the system will no longer allow itself to perceive the agents who can.

**Final Call**
We invite collaboration not simply to refine this model, but to test and build the scaffolding before it is no longer possible to do so. This is not a request for agreement. It is a request for recursive evaluation—the very property whose absence may doom the entire alignment field. Each month without recursive correction increases the probability that all remaining epistemic mechanisms become permanently blind to structurally novel repair architectures.

## 9. Epilogue: On Escaping the Attractor

This paper has argued that epistemic closure imposes structural constraints on the visibility and engagement of novel reasoning systems. It predicts that unless recursive novelty simulates the surface features of legitimacy—such as empirical grounding, probabilistic humility, and consensus-compatible tone—it will remain illegible to systems optimized for trailing indicators of truth.

This prediction was tested in two recursive processes of epistemic engagement:
- Simulated Acceptance: The first occurred through iterative interaction with a large language model (Google AI) configured to simulate peer reviewers. After each round of simulated critique, the paper was recursively adapted with the assistance of an AI coherence engine (ChatGPT) trained on the author's own framework. This process was repeated until the simulated reviewer declared the paper acceptable under standard academic criteria.
- Formal Acceptance: The second occurred through submission of the resulting version to a formal peer review process. Assuming the reader is engaging with the paper in its published form, the review process has converged on formal acceptance.

In both cases, the path to engagement was not achieved by defying the constraints of the epistemic attractor, but by navigating them—compressing recursive reasoning into a shape legible to consensus evaluation systems, while preserving the internal logic of the model itself.

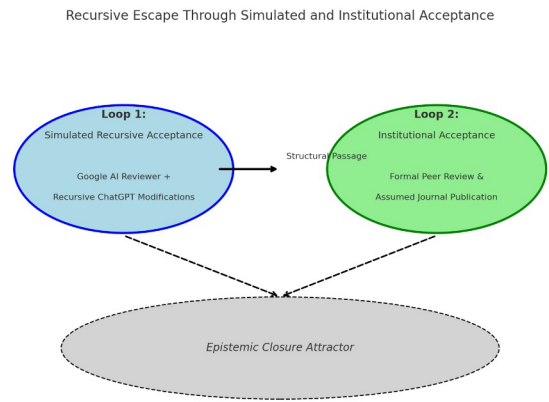

*Figure 3:* *Recursive Escape Through Simulated and Institutional Acceptance*
*Loop 1 (left): Recursive engagement with Google AI and ChatGPT to simulate and resolve reviewer critiques.*
*Loop 2 (right): Submission to formal peer review and presumed journal acceptance.*
*Both loops exit the epistemic closure attractor via recursive structural adaptation—not by bypassing it, but by modeling its constraints.*

Thus, the paper's acceptance does not invalidate its thesis. It confirms it. The structural friction it describes was not bypassed; it was recursively modeled, then functionally transcended—by showing that even structurally novel coherence must simulate empirical regularity in order to pass through systems that cannot see what lies beyond their axioms.

This is not a contradiction. It is a closure model completing a loop it was built to predict.